\documentclass{article}
\usepackage{spconf,amsmath,graphicx}

\usepackage{cleveref} 
\usepackage[table]{xcolor} 
\usepackage{pifont} 
\newcommand{\cmark}{\ding{51}} 

\usepackage{url}

\title{LeMoRe: Learn More Details for Lightweight Semantic Segmentation}

\name{Mian Muhammad Naeem Abid, Nancy Mehta, Zongwei Wu, Radu Timofte\thanks{This work was supported by the Alexander von Humboldt Foundation.}}
\address{Computer Vision Lab, CAIDAS \& IFI, University of Würzburg, Germany}

\usepackage{etoolbox}


\usepackage{wrapfig}
\usepackage{multirow}

\begin{document}

\maketitle

\begin{abstract}
Lightweight semantic segmentation is essential for many downstream vision tasks. Unfortunately, existing methods often struggle to balance efficiency and performance due to the complexity of feature modeling. Many of these existing approaches are constrained by rigid architectures and implicit representation learning, often characterized by parameter-heavy designs and a reliance on computationally intensive Vision Transformer-based frameworks. In this work, we introduce an efficient paradigm by synergizing explicit and implicit modeling to balance computational efficiency with representational fidelity. Our method combines well-defined Cartesian directions with explicitly modeled views and implicitly inferred intermediate representations, efficiently capturing global dependencies through a  nested attention mechanism. Extensive experiments on challenging datasets, including ADE20K, CityScapes, Pascal Context, and COCO-Stuff, demonstrate that LeMoRe strikes an effective balance between performance and efficiency. \url{https://github.com/miannaeem-lab/LeMoRe}
\end{abstract}

\section{Introduction}   

\begin{figure}[!ht]
  \centering
  \includegraphics[width=.8\linewidth]{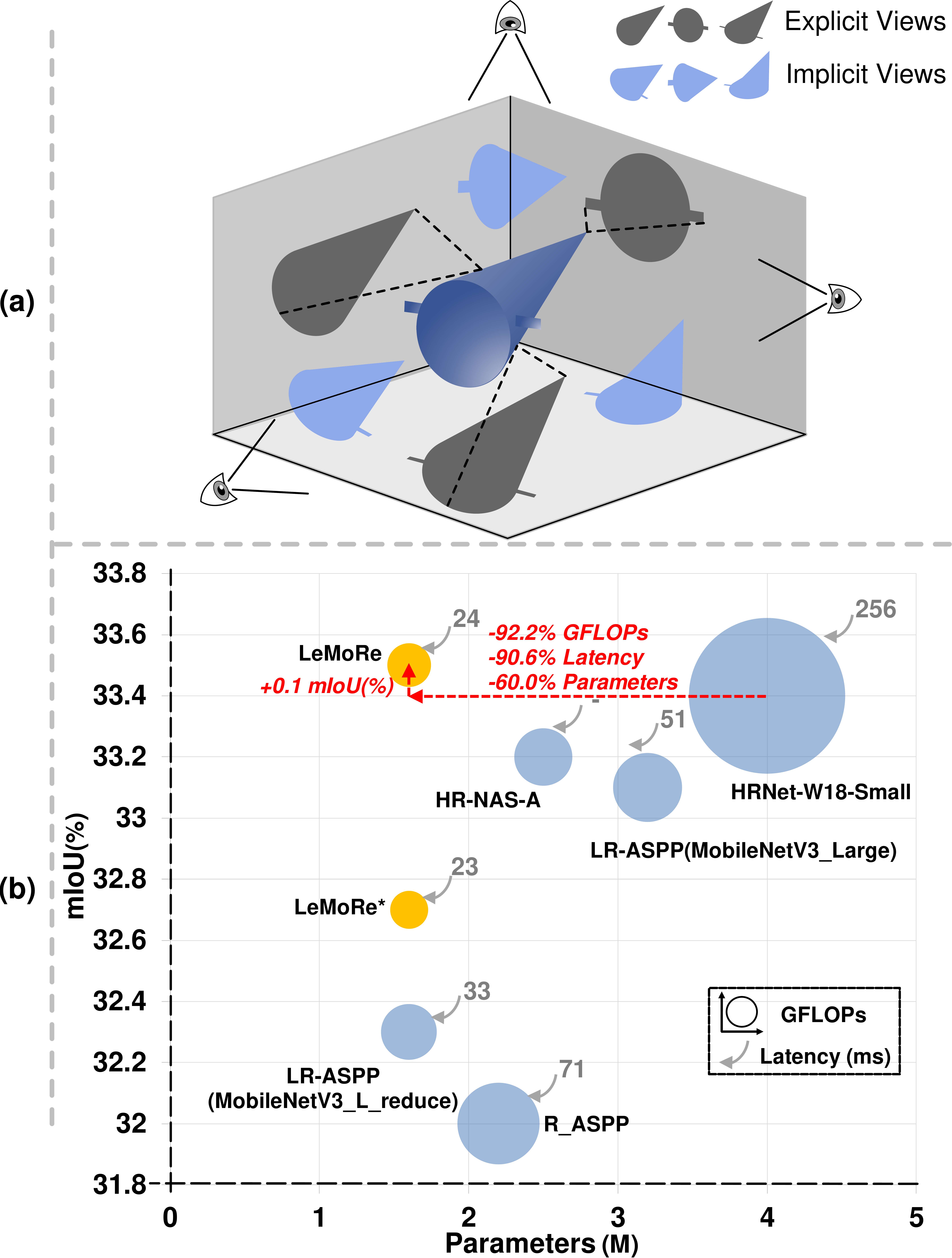}
    \vspace{-4mm}
  \caption{
  (a) We propose replacing costly feature modeling with interpretable projections into lower-dimensional spaces for improved efficiency. These projections or views capture various feature aspects. Explicit views align with predefined directions like Cartesian coordinates, while implicit views represent intermediate positions learned through nested attention for global dependencies.
  (b) Our method performs better than counterpart solutions with less computational cost (\Cref{sec:results_ade20K}). 
  }
  \label{fig:different_views_and_comparison_of_all_models}
  \vspace{-5mm}
\end{figure}

Semantic segmentation is the task of assigning class labels to each pixel in an image, playing a crucial role in applications such as autonomous driving and AR/VR \cite{weaver2022systematic}.
Traditional approaches use Convolutional Neural Networks (CNNs) for segmentation based on local dependencies. Recently, Vision Transformers (ViTs) have dominated this field by leveraging global feature dependencies within images \cite{wang2022unetformer,yeom2023u}. 

Despite their exceptional performance, ViTs face significant challenges, including quadratic computational cost and limited local awareness. Their reliance on numerous learnable parameters makes them unsuitable for resource-constrained environments. Efficient approaches that combine CNNs and ViTs attempt to address these issues but often fall short, struggling to balance performance and efficiency, as shown in ~\Cref{fig:different_views_and_comparison_of_all_models}(b). 
One underlying limitation is their reliance on single-view representations, which fail to fully capture the diverse spatial and contextual relationships within features.
To address these limitations, we propose a novel framework that synergistically combines explicit and implicit views in a decomposition-modeling-reconstruction paradigm  to achieve efficient and comprehensive feature modeling. By decomposing features into localized yet complementary explicit views, as shown in ~\Cref{fig:different_views_and_comparison_of_all_models}(a), our method captures diverse aspects of the data while significantly reducing computational overhead (\Cref{fig:different_views_and_comparison_of_all_models}(b)). The final feature representation is then reconstructed by aggregating these views, effectively replicating the robustness of conventional methods while enhancing processing efficiency. 

Technically, we model the feature through distinct explicit views aligned with Cartesian directions, projecting the feature map into a multidimensional, orthogonal basis. This is achieved via Cartesian encoder, where three 1$\times$1 convolutions are applied along different directions of the feature map, with each convolution independently capturing specific components. This explicit modeling not only reduces computational and memory requirements, but also maximizes joint entropy, allowing for higher-order local relationships to be effectively captured. To further enrich the feature representation, we introduce implicitly learned virtual views. Unlike the predefined Cartesian directions, these views are dynamically modeled through nested transformer attention over query-key pairs. This implicit modeling enhances the system's ability to capture complex global dependencies while maintaining low computational costs.

We further validate the efficacy of LeMoRe through extensive experiments on several challenging datasets.
Our results demonstrate that LeMoRe achieves a compelling trade-off between segmentation accuracy and computational efficiency, making it well-suited for applications that demand both high performance and low resource consumption.

To conclude, our contributions are as follows: 
\begin{itemize} 
\vspace{-1mm}
\item We introduce LeMoRe, a novel semantic segmentation framework that balances segmentation accuracy and computational efficiency by modeling features through distinct Cartesian and learned virtual views. 
\vspace{-1.7mm}
\item We propose a multiview feature modeling approach that reduces computational and memory costs by projecting features into independent components and capturing higher-order local relationships with minimal overhead. 
\vspace{-1.7mm}
\item We rigorously evaluate the effectiveness of LeMoRe through extensive experiments on challenging datasets, such as  ADE20K \cite{zhou2017scene}, CityScapes \cite{cordts2016cityscapes}, Pascal Context \cite{mottaghi2014role}, and COCO-Stuff \cite{caesar2018coco}, showing its strong performance in resource-constrained environments. \end{itemize}

\vspace{-3mm}

\section{Related Work} 
\label{sec:related_work}

\noindent \textbf{Efficient Convolutional Neural Networks:}
Convolutional Neural Networks (CNNs) have significantly advanced computer vision \cite{he2016deep}. However, their high computational demands make them unsuitable for resource-limited environments. Efficient CNN architectures like MobileNet \cite{howard2017mobilenets, sandler2018mobilenetv2}, ShuffleNet \cite{ma2018shufflenet}, and GhostNet \cite{han2020ghostnet} address these challenges with techniques such as inverted bottlenecks and depth-wise convolutions. 
Unlike these efficient CNNs, which primarily focus on local feature extraction, LeMoRe captures both global and local context by leveraging cross-dimensional interactions and multi-scale representations through explicit and implicit views.

\noindent \textbf{Lightweight Vision Transformers:}
Vision Transformers (ViTs) like ViT \cite{dosovitskiy2020image} have revolutionized image recognition tasks but face challenges in resource-constrained environments due to their high computational complexity. 
Enhancements such as DeiT \cite{touvron2021training} and LeViT \cite{graham2021levit} have improved self-attention efficiency and incorporated spatial inductive biases. Lightweight ViTs like EfficientFormer \cite{li2022efficientformer} and MobileViT \cite{mehta2021mobilevit} merge the strengths of CNNs and transformers, featuring efficient attention mechanisms and compact designs. Unlike lightweight ViTs, LeMoRe employs a dual approach combining explicit views through Cartesian Encoder with implicit views via Nested Attention, enhancing computational and memory efficiency while effectively capturing global dependencies.

\vspace{-3mm}

\section{Proposed Method}
\label{proposed_method}
\begin{figure*}[t]
  \centering
  \includegraphics[width=\linewidth]{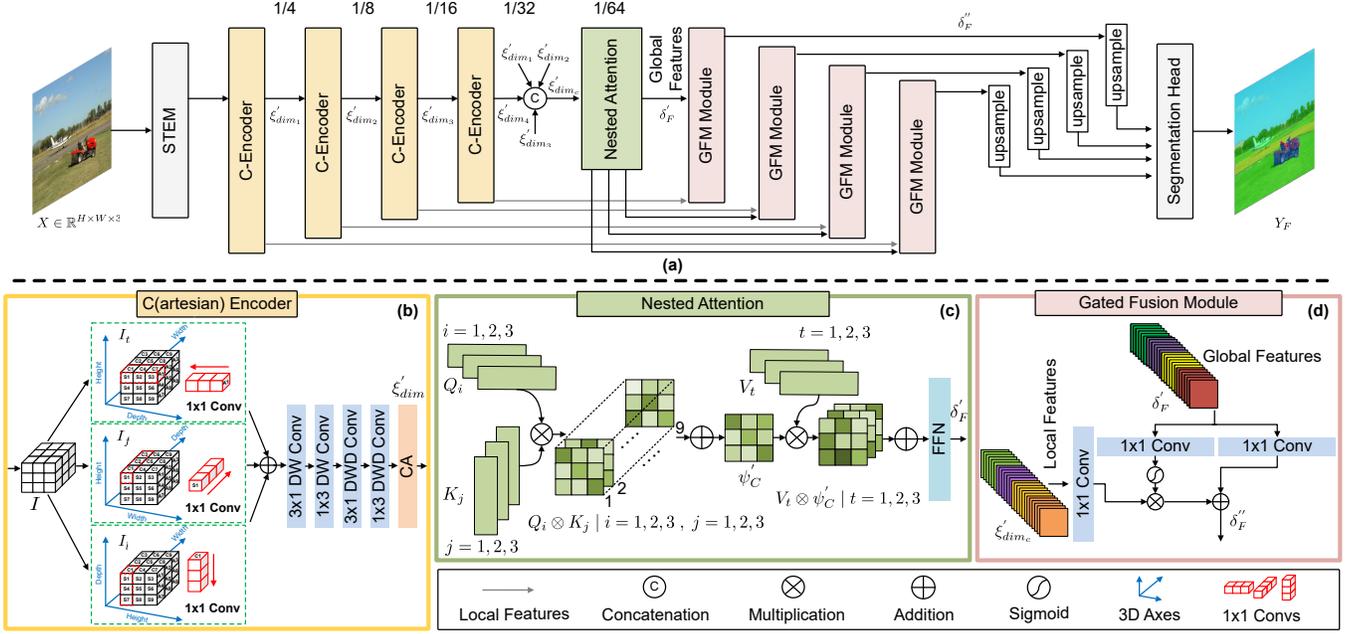}
  \caption{ 
  \textit{Architecture Overview}: (a) shows the proposed architecture of LeMoRe, highlighting how our method balances efficiency and performance through multiview modeling. (b) depicts the Cartesian Encoder, which enhances contextual understanding and feature richness by extracting explicit views in three dimensions using well-defined Cartesian directions. (c) illustrates the Nested Attention mechanism, which enriches the feature modeling by learning complex relationships within the data through implicit views, maintaining low computational cost via efficient attention over each query-key pair. (d) presents the GFM module, which dynamically fuses global and local features to improve segmentation performance.
  }
  \label{fig:Full_Proposed_Method_Diagram_big_fonts}
  \vspace{-3mm}
\end{figure*}

\vspace{-3mm}
As illustrated in \Cref{fig:Full_Proposed_Method_Diagram_big_fonts}, the proposed architecture comprises of different parts: Cartesian Encoder for efficiently encoding the explicit information, Nested Attention in the bottleneck for effective detail coverage via learning implicit information, and Gated Fusion Module for efficient complementary feature integration. 
\vspace{-3.5mm}

\subsection{Cartesian Encoder}
\label{sec:tape}
LeMoRe employs Cartesian Encoders (C-Encoder) in a hierarchical fashion to structure feature representations into well-defined explicit views, enabling more efficient and interpretable feature encoding. This approach replaces the conventional encoder \cite{kirillov2019panoptic}, which is computationally expensive, with a structured orthogonal decomposition strategy that enhances interpretability, while significantly reducing computational overhead.

The projection process utilizes $1 \times 1$ convolutions applied along distinct spatial orientations of the feature map, ensuring that each convolution isolates independent, orthogonal components. By decomposing the feature space into structured subspaces and subsequently regrouping them, the model enhances both representational efficiency and feature discrimination.  Formally, given a feature map $I$, the C-Encoder hierarchically extracts spatial features across three principal dimensions, generating Transverse $t$, Frontal $f$, and Lateral $l$ views. The overall process is formulated as:
\begin{equation}
\begin{split}
    & I_{t}^{}, I_{f}^{}, I_{l}^{} = \rho_{t}(I), \rho_{f}(I), \rho_{l}(I) \\
    & \xi_{dim} = \zeta_{t}^{1 \times 1}(I_{t}) + \zeta_{f}^{1 \times 1}(I_{f}) + \zeta_{l}^{1 \times 1}(I_{l})
\end{split}
  \label{eq:2}
\end{equation}
where $\rho_{x}(I)$ are spatial permutation along one direction $x$ in $[t, f, l]$. $\zeta_{x}^{1 \times 1}$ are the processing convolutions. After the decomposition-modeling-regrouping, we apply a more in-depth yet lightweight refinement through cascaded depthwise convolutions (DW Conv), dilated depthwise convolutions (DWD Conv), and Channel Attention (CA) to obtain $\xi'_{dim}$ as shown in \Cref{fig:Full_Proposed_Method_Diagram_big_fonts}(b).

\subsection{Nested Attention}
\label{sec:tsca}
In addition to the explicitly defined Cartesian directions, we introduce implicit view learning through nested transformer attention over query-key pairs. This enhances global feature interaction complementing the localized encoding of the C-Encoder while preserving computational efficiency. The Nested Attention generates several sets - three in our case, of query ($Q$), key ($K$), and value ($V$) pairs using $1 \times 1$ convolutions:
\begin{equation}
(Q_e, K_e, V_e) = \zeta^{1 \times 1}(\xi_{dim_c}^{'}) \quad | \quad e = 1, 2, 3
  \label{eq:tri_dim}
\end{equation}

For each query $Q_e$, attention maps are computed against all keys $(K_1, K_2, K_3)$, and the resulting interactions are aggregated into $\psi_{C}^{'}$:

\begin{equation}
\psi_{C}^{'} = \sum_{i=1}^{3} \sum_{j=1}^{3} Q_i \otimes K_j 
  \label{eq:7}
\end{equation}

This aggregated attention map is then used to weight the corresponding value pairs $(V_1, V_2, V_3)$, followed by processing through a feed-forward network (FFN) as shown:
\begin{equation}
\delta_{F}^{'} = FFN(\sum_{t=1}^{3} \psi_{C}^{'} \otimes V_t)
  \label{eq:9}
\end{equation}
Unlike conventional attention, we leverage a structured decomposition and regrouping strategy to achieve efficient and effective feature modeling, contributing to our lightweight model.

\subsection{Gated Fusion Module and Segmentation Head}
\label{sec:gfm_sh}

For the decoder, we adopt a gating fusion mechanism that integrates local features from the C-Encoder ($\xi_{dim_c}^{'}$), representing explicit views, and global features from the Nested Attention ($\delta_{F}^{'}$), representing implicit views. The objective is to selectively combine these complementary aspects, which are equally important for segmentation tasks, to achieve precise and robust feature decoding. This gating integration is achieved using simple $1 \times 1$ convolutions and a sigmoid activation ($\sigma$) as:
\begin{equation}
\delta_{F}^{''} = (\zeta^{1 \times 1}(\xi_{dim_c}^{'}) \bar{\otimes} (\sigma(\zeta^{1 \times 1}(\delta_{F}^{'})))) + \zeta^{1 \times 1}(\delta_{F}^{'})
  \label{eq:10}
\end{equation}
Finally, the corresponding enriched upsampled features are forwarded to the Segmentation Head with two $1 \times 1$ convolutional layers to produce the final segmentation map, $Y_{F}$.

\vspace{-2mm}

\section{Experiments}
\label{sec:experiments}

\begin{table*}[t]
\footnotesize
  \caption{ Results of various models on ADE20K validation set~\cite{zhou2017scene}. * indicates results from models trained with $448 \times 448$ input size.  
  GFLOPs are reported for input resolution of $512 \times 512$. Single-scale inference is used to report mIoU.
  }
  \label{tab:results_ade20k}
  \centering
  \begin{tabular}{@{}c|c|c|c|c|c|c@{}}
    \hline
    \multirow{14}{*}{\rotatebox{90}{Heavyweight Models}} & Models & Encoder & mIoU & GFLOPs & Parameters & Latency(ms) \\
    \cline{2-7}
    & PSPNet \cite{zhao2017pyramid} & MobileNetV2~\cite{sandler2018mobilenetv2} & 29.6 & 52.2 & 13.7M & 426 \\
    & FCN-8s \cite{long2015fully} & MobileNetV2~\cite{sandler2018mobilenetv2} & 19.7 & 39.6 & 9.8M & 406 \\
    & Semantic FPN \cite{kirillov2019panoptic} & ConvMLP-S~\cite{li2023convmlp} & 35.8 & 33.8 & 12.8M & 311 \\
    & DeepLabV3+ \cite{chen2018encoder} & EfficientNet~\cite{tan2019efficientnet} & 36.2 & 26.9 & 17.1M & 388 \\
    & DeepLabV3+ \cite{chen2018encoder} & MobileNetV2~\cite{sandler2018mobilenetv2} & 38.1 & 25.8 & 15.4M & 414 \\
    & Lite-ASPP \cite{chen2018encoder} & ResNet18~\cite{he2016deep} & 37.5 & 19.2 & 12.5M & 259 \\
    & PEM~\cite{cavagnero2024pem} & STDC1~\cite{cavagnero2024pem} & 39.6 & 16.0 & 17.0M & - \\
    & DeepLabV3+ \cite{chen2018encoder} & ShuffleNetv2-1.5x~\cite{ma2018shufflenet} & 37.6 & 15.3 & 16.9M & 384 \\
    & HRNet-Small \cite{yuan2020object} & HRNet-W18-Small~\cite{yuan2020object} & 33.4 & 10.2 & 4.0M & 256 \\
    & SegFormer \cite{xie2021segformer} & MiT-B0~\cite{xie2021segformer} & 37.4 & 8.4 & 3.8M & 308 \\
    & FeedFormer-B0 \cite{shim2023feedformer} & MiT-B0~\cite{xie2021segformer} & 39.2 & 7.8 & 4.5M & - \\
    & SegNeXt \cite{guo2022segnext} & SegNeXt-T~\cite{guo2022segnext} & 41.1 & 6.6 & 4.3M & - \\
    & U-MixFormer \cite{yeom2023u} & MiT-B0~\cite{xie2021segformer} & 41.2 & 6.1 & 6.1M & - \\
    \hline
    \multirow{8}{*}{\rotatebox{90}{Lightweight Models}} & Lite-ASPP \cite{chen2018encoder} & MobileNetV2~\cite{sandler2018mobilenetv2} & 36.6 & 4.4 & 2.9M & 94 \\
    & R-ASPP \cite{sandler2018mobilenetv2} & MobileNetV2~\cite{sandler2018mobilenetv2} & 32.0 & 2.8 & 2.2M & 71 \\
    & HR-NAS-B \cite{ding2021hr} & Searched~\cite{ding2021hr} & 34.9 & 2.2 & 3.9M & - \\
    & LR-ASPP \cite{howard2019searching} & MobileNetV3-Large~\cite{howard2019searching} & 33.1 & 2.0 & 3.2M & 51 \\
    & HR-NAS-A \cite{ding2021hr} & Searched~\cite{ding2021hr} & 33.2 & 1.4 & 2.5M & - \\
    & LR-ASPP \cite{howard2019searching} & MobileNetV3-Large-reduce~\cite{howard2019searching} & 32.3 & 1.3 & 1.6M & 33 \\
    \cline{2-7}
    & \cellcolor{gray!20} LeMoRe* (Ours) & \cellcolor{gray!20} Ours & \cellcolor{gray!20}32.7 & \cellcolor{gray!20}0.6 & \cellcolor{gray!20}1.6M & \cellcolor{gray!20}23 \\
    & \cellcolor{gray!20} LeMoRe (Ours) & \cellcolor{gray!20} Ours & \cellcolor{gray!20}33.5 & \cellcolor{gray!20}0.8 & \cellcolor{gray!20}1.6M & \cellcolor{gray!20}24 \\
    \hline
  \end{tabular}
\vspace{-3mm}
\end{table*}

\subsection{Results on ADE20K} 
\label{sec:results_ade20K}

In \Cref{tab:results_ade20k}, we compared the performance of our LeMoRe on the ADE20K dataset with other state-of-the-art methods. 
Compared to the best-performing U-Mixformer, which uses the lightweight MiT-B0 encoder, our method achieves highly competitive performance with 73.8\% less parameters and a GFLOPs reduction of 86.9\%. 
This efficiency is primarily driven by our hybrid network design, particularly the Cartesian Encoder, which relies on lightweight convolutions rather than the pure transformer attention of MiT backbones, ensuring significantly lower computational load.
Similarly, compared to the lightweight LR-ASPP solution, our LeMoRe achieves a performance improvement of +1.0\%, while reducing computational cost by 38.5\%. This is attributed to our Nested Attention mechanism, which efficiently captures global context to complement local details -- an aspect missing in LR-ASPP's MobileNet-based design. In nutshell, our method stands out against state-of-the-art solutions by providing a balanced trade-off between performance and efficiency. 
LeMoRe achieves notable reductions in GFLOPs and latency, setting a new benchmark for future lightweight models in semantic segmentation.
\textit{Details of datasets, measures, and implementation details are provided in supp. material.}\footnote{https://sigport.org/documents/supplementary-material-lemore}

\begin{figure}[t]
  \centering
  \includegraphics[width=\linewidth]{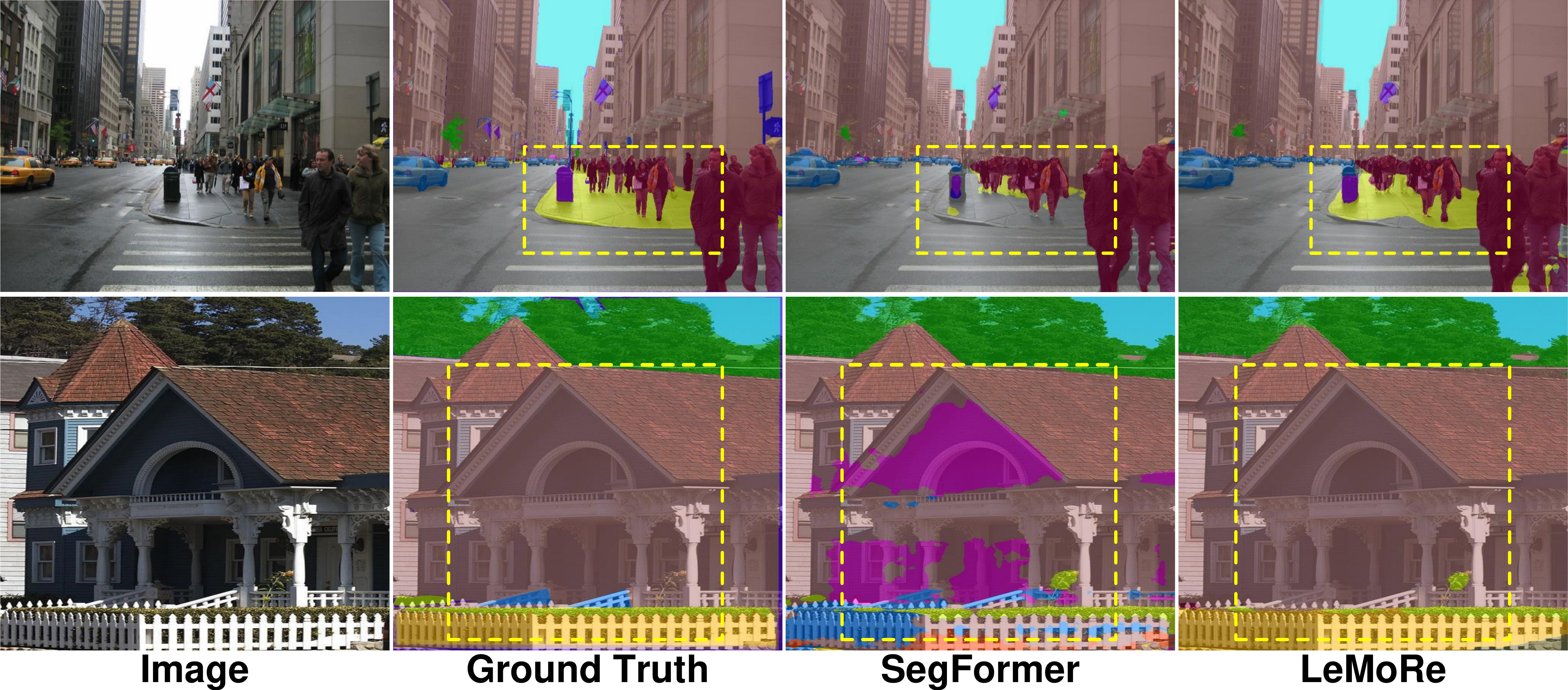}
  \vspace{-6mm}
  \caption{The visualization of Image, Ground Truth, SegFormer and LeMoRe results on the ADE20K validation set. The results highlight the proposed model's effectiveness in producing high-quality segmentation maps with improved spatial consistency.
  }
  \label{fig:proposed_method_visual_results}
  \vspace{-4mm}
\end{figure}

\textbf{Visual Results:}
~\Cref{fig:proposed_method_visual_results} presents the comparative visual results of our proposed LeMoRe model on the validation set of the ADE20K benchmark dataset. Here, we compare against original images, ground truth, and SegFormer \cite{xie2021segformer} results. 
In the first example, we observe that our model produces more complete pedestrian regions. This is because the pedestrian region has similar textural clues as the road, thus sharing similar pixel values. Therefore, it is difficult for SegFormer which lacks the locality processing ability. In contrast, our Cartesian encoder incorporates local modeling, enabling the detection of fine-grained details and effectively distinguishing between pedestrian regions and the road, resulting in better segmentation. From the second example, we can observe that SegFormer is sensitive to light changes, leading to misclassification and separation of local regions which are naturally sharing the same class. Differently, thanks to our hybrid design, we can effectively merge the local awareness with the global dependencies through our joint modeling and gating fusion, achieving to more robust segmentation despite the shadows.

\vspace{-3mm}
\subsubsection{Ablation Study}

\noindent \textbf{Impact of the Cartesian Encoder:}
The Cartesian Encoder Block enhances segmentation performance by explicitly modeling the multi-dimensional spatial views and incorporating Channel Attention (CA). As shown in ~\Cref{tab:results_ade20K_ablation}, starting with the Transverse view, the baseline model achieves an mIoU of 27.2\% with 0.68 GFLOPs and 1.12M parameters. Explicitly adding the Frontal and Lateral views progressively increases mIoU to 27.6\% (+0.4\%) and 29.1\% (+1.5\%), demonstrating the benefit of capturing richer spatial context across orthogonal dimensions.
Integrating CA with all three explicit views further improves mIoU to 29.6\% (+0.5\%) while maintaining efficiency at 0.79 GFLOPs and 1.22M parameters. The CA block adaptively emphasizes key features across the views, enabling finer detail capture and superior segmentation performance.

\noindent \textbf{Contribution of Nested Attention Mechanism:} Integrating the Nested Attention (N-Attn) mechanism significantly enhances segmentation by refining feature relevance and suppressing noise. With N-Attn and only the Transverse view, mIoU improved to 32.2\% (+5.0\%). Adding the Frontal view raised mIoU to 32.7\% at 0.76 GFLOPs and 1.52M parameters, while incorporating the Lateral view further increased it to 33.3\% (+0.6\%). The highest mIoU, 33.5\%, was achieved by combining N-Attn with Channel Attention (CA) and all three Cartesian dimensions at 0.81 GFLOPs and 1.60M parameters. This demonstrates the synergy of N-Attn and multi-dimensional fusion in enhancing segmentation precision with computational efficiency.
\begin{table}[t] 
\footnotesize
  \caption{Ablation studies of LeMoRe model on the ADE20K validation set~\cite{zhou2017scene}. 
  }
  \label{tab:results_ade20K_ablation}
  \centering
  \resizebox{\columnwidth}{!}{%
  \begin{tabular}{@{}c|c|c|c|c|c|c|c@{}}
    \hline
    \multicolumn{4}{c|}{C-Encoder}  & N-Attn & mIoU & GFLOPs & Parameters\\
    \cline{1-4}
    \multicolumn{3}{c|}{Views} & CA &  & &  & \\
    \cline{1-3}
    Transverse & Frontal & Lateral &  &  &  &  & \\
    \hline
    \cmark & & & & & 27.2 & 0.68 & 1.12M \\
    \cmark & \cmark & & & & 27.6 & 0.74 & 1.14M\\
    \cmark & \cmark & \cmark & & & 29.1 & 0.78 & 1.16M  \\
    \cmark & \cmark & \cmark & \cmark & & 29.6 & 0.79 & 1.22M \\
    \cmark & & &  & \cmark & 32.2 & 0.70 & 1.50M \\
    \cmark & \cmark & &  & \cmark & 32.7 & 0.76 & 1.52M \\
    \cmark & \cmark & \cmark &  & \cmark & 33.3 & 0.80 & 1.54M \\
    \rowcolor{gray!20}
    \cmark & \cmark & \cmark & \cmark & \cmark & 33.5 & 0.81 & 1.60M \\
 \hline
    \end{tabular}
    }
    \vspace{-4mm}
\end{table}

\begin{table}[t]
\footnotesize
  \caption{Results on CityScapes validation set~\cite{cordts2016cityscapes}.
  }
  \label{tab:results_cityscapes}
  \centering
  \resizebox{\columnwidth}{!}{%
  \begin{tabular}{@{}c|c|c|c@{}}
    \hline
    Methods & Encoder & mIoU & GFLOPs \\
    \hline
    PSPNet~\cite{zhao2017pyramid} & MobileNetV2~\cite{sandler2018mobilenetv2} & 70.2 & 423.4 \\
    FCN~\cite{long2015fully} & MobileNetV2~\cite{sandler2018mobilenetv2} & 61.5 & 317.1 \\
    SegFormer~\cite{xie2021segformer} & MiT-B0~\cite{xie2021segformer} & 71.9 & 17.7 \\
    L-ASPP~\cite{chen2018encoder} & MobileNetV2~\cite{sandler2018mobilenetv2} & 72.7 & 12.6 \\
    LR-ASPP~\cite{howard2019searching} & MobileNetV3-Large~\cite{howard2019searching} & 72.4 & 9.7 \\
    LR-ASPP~\cite{howard2019searching} & MobileNetV3-Small~\cite{howard2019searching} & 68.4 & 2.9 \\
    \rowcolor{gray!20}
    LeMoRe (Ours) & Ours & 65.0 & 1.7\\
      \hline
    \end{tabular}
    }
    \vspace{-4mm}
\end{table}
\begin{table}[t]
\footnotesize
  \caption{Results on PASCAL Context test set~\cite{mottaghi2014role}.
  }
  \label{tab:results_pascal_context}
  \centering
  \resizebox{\columnwidth}{!}{%
  \begin{tabular}{@{}c|c|c|c|c@{}}
    \hline
    Methods & Backbone & mIoU$^{59}$ & mIoU$^{60}$ & GFLOPs \\
    \hline
    DeepLabV3+~\cite{chen2018encoder} & ENet-s16~\cite{paszke2016enet} & 43.07 & 39.19 & 23.00 \\
    DeepLabV3+~\cite{chen2018encoder} & MobileNetV2-s16~\cite{sandler2018mobilenetv2} & 42.34 & 38.59 & 22.24 \\
    LR-ASPP~\cite{howard2019searching} & MobileNetV3-s16~\cite{howard2019searching} & 38.02 & 35.05 & 2.04 \\
    \rowcolor{gray!20}
    LeMoRe (Ours) & Ours & 35.76 & 31.48 & 0.68\\
      \hline
    \end{tabular}
    }
    \vspace{-4mm}
\end{table}
\begin{table}[t]
\footnotesize
  \caption{Results on COCO-Stuff test set~\cite{caesar2018coco}.
  }
  \label{tab:results_coco_stuff}
  \centering
  \resizebox{\columnwidth}{!}{%
  \begin{tabular}{@{}c|c|c|c@{}}
    \hline
    Methods & Encoder & mIoU & GFLOPs \\
    \hline
    PSPNet~\cite{zhao2017pyramid} & MobileNetV2-s8~\cite{sandler2018mobilenetv2} & 30.14 & 52.94 \\
    DeepLabV3+~\cite{chen2018encoder} & EfficientNet-s16~\cite{tan2019efficientnet} & 31.45 & 27.10 \\
    DeepLabV3+~\cite{chen2018encoder} & MobileNetV2-s16~\cite{sandler2018mobilenetv2} & 29.88 & 25.90 \\
    LR-ASPP~\cite{howard2019searching} & MobileNetV3-s16~\cite{howard2019searching} & 25.16 & 2.37 \\
    \rowcolor{gray!20}
    LeMoRe (Ours) & Ours & 27.19 & 0.8 \\
      \hline
    \end{tabular}
    }
    \vspace{-4mm}
\end{table}
\vspace{-3mm}

\subsection{Results on CityScapes, PASCAL-Context, and COCO-Stuff}
\begin{table}[ht!]
\footnotesize
  \caption{Results of object detection based on RetinaNet.
  }
  \label{tab:results_object_detection}
  \centering
  \begin{tabular}{@{}c|c|c|c@{}}
    \hline
    Backbone & mAP & GFLOPs & Parameters \\
    \hline
    ShuffleNetV2~\cite{ma2018shufflenet} & 25.9 & 161 & 10.4M  \\
    MobileNetV3~\cite{howard2019searching} & 27.2 & 162 & 12.3M  \\
    \rowcolor{gray!20}
    LeMoRe (Ours) & 29.3 & 161 & 10.9M  \\
      \hline
    \end{tabular}
    \vspace{-4mm}
\end{table}

\vspace{-3mm}
As shown in \Cref{tab:results_cityscapes}, \Cref{tab:results_pascal_context}, and  \Cref{tab:results_coco_stuff}, LeMoRe demonstrates superior computational efficiency and competitive performance across diverse datasets, owing to its innovative Cartesian Encoder and Nested Attention mechanism. On Cityscapes, LeMoRe achieves an mIoU of 65.0\% with only 1.7 GFLOPs, representing up to a 99.60\% reduction in computational cost compared to PSPNet with MobileNetV2, while maintaining real-time suitability for high-resolution urban scenes. On PASCAL Context, LeMoRe delivers an mIoU$^{59}$ of 35.76\% and an mIoU$^{60}$ of 31.48\% with just 0.68 GFLOPs—97.04\% lower than DeepLabV3+ with ENet-s16, showcasing its ability to handle intricate scenes and diverse classes with minimal computational demand. Similarly, on COCO-Stuff, LeMoRe achieves 27.19\% mIoU with 0.8 GFLOPs, outperforming LR-ASPP with MobileNetV3-Small while reducing GFLOPs by 66.25\%. The underperformance of LR-ASPP can be attributed to its lack of explicit multi-dimensional modeling and limited global context awareness, which restricts its ability to capture intricate spatial features. In contrast, LeMoRe's Cartesian Encoder explicitly models the multi-dimensional views for richer spatial local feature extraction, while the Nested Attention mechanism efficiently captures global context. This synergy ensures a remarkable balance between segmentation accuracy and computational efficiency, making LeMoRe ideal for resource-constrained, real-time applications.

\vspace{-2mm}
\subsection{Object Detection}
We evaluated our LeMoRe model's robustness on the object detection task. As shown in \Cref{tab:results_object_detection}, the results, LeMoRe exhibits enhanced generalization capability and superior performance.

\vspace{-4mm}
\section{Conclusion}
\vspace{-2mm}
In this paper, we introduced LeMoRe, a novel lightweight and efficient architecture for semantic segmentation. LeMoRe combines explicit views through Cartesian coordinates with implicit views learned via nested transformers, achieving comprehensive and efficient feature representation. This dual approach enhances computational and memory efficiency while capturing local-global dependencies.  
Our experiments on challenging datasets show that LeMoRe balances segmentation accuracy and computational efficiency. 
These results highlight LeMoRe's potential for real-world applications requiring high performance and low resource consumption. Future work will explore further optimizations and applications of LeMoRe in different domains, advancing semantic segmentation technology.

\bibliographystyle{IEEEbib}
\bibliography{strings,refs}

\onecolumn
\newpage
\twocolumn


\section*{\centering --- Supplementary Material ---}

\subsection*{ImageNet Pre-training} 
To ensure a fair comparison, we initialize the LeMoRe model with pre-trained parameters from ImageNet. The classification head of LeMoRe includes an average pooling layer followed by a linear layer, leveraging global semantic representations to generate class scores. Given the low resolution of the input images ($224 \times 224$), the target resolution of the input tokens for Nested Attention is set to $\frac{1}{64} \times \frac{1}{64}$ of the input dimensions.
Quantitative results of the proposed LeMoRe model on the ImageNet-1K dataset are shown in~\Cref{tab:results_imagenet1k}.
\vspace{-3mm}
\begin{table}[h]
  \caption{LeMoRe results for ImageNet classification.
  }
  \label{tab:results_imagenet1k}
  \small
  \centering
  \begin{tabular}{@{}c|c|c|c@{}}
    \hline
    Method & Input Size & Top-1 Accuracy(\%) & Parameters \\
    \hline
    LeMoRe & $224\times224$ & 64.1 & 1.67M \\
      \hline
    \end{tabular}
\end{table}

\subsection*{Datasets and Measures}

For semantic segmentation, experiments are conducted on four benchmark datasets: ADE20K \cite{zhou2017scene}, CityScapes \cite{cordts2016cityscapes}, PASCAL Context \cite{mottaghi2014role}, and COCO-Stuff \cite{caesar2018coco}. The \textbf{ADE20K}~\cite{zhou2017scene} dataset consists of 25,000 images spanning 150 class categories, with 20,000 images for training, 2,000 for validation, and 3,000 for testing. The \textbf{CityScapes}~\cite{cordts2016cityscapes} dataset, containing 19 fine-class annotations, comprises 2,975 images for training and 500 images for validation and testing. The \textbf{PASCAL Context}~\cite{mottaghi2014role} dataset includes 10,103 images, featuring 1 background and 59 semantic labels, divided into 4,998 training images and 5,105 testing images. \textbf{COCO-Stuff}~\cite{caesar2018coco}, derived from pixel-level annotations on the COCO dataset, includes 10,000 images, with 9,000 for training and 1,000 for testing.

In line with recent literature \cite{kang2024metaseg, cavagnero2024pem, zhang2022topformer}, we report results using standard metrics: Mean Intersection over Union (mIoU) for segmentation accuracy, Giga Floating Point Operations per Second (GFLOPs), latency, and the number of parameters.

\subsection*{Implementation Details}

Our implementation is based on the PyTorch framework and the MMSegmentation toolbox \cite{mmseg2020}. All models, including our proposed LeMoRe are initially pretrained on the ImageNet-1K dataset \cite{deng2009imagenet} before being fine-tuned on semantic segmentation datasets. Batch Normalization layers are applied after almost every convolution layer, except the final output layer. For the ADE20K dataset, we adopt data augmentations similar to those in \cite{xie2021segformer} to ensure fair comparison. We use a batch size of 16 and follow a 160K scheduler as described in \cite{xie2021segformer}. Applied augmentations include random scaling, cropping, horizontal flipping, and resizing. For the CityScapes dataset, we apply the same data augmentations as in \cite{xie2021segformer}, with images resized and rescaled to a crop size of $1024 \times 512$. Across all datasets, we set an initial learning rate of $1.2 \times 10^{-4}$ with a weight decay value of $0.01$. For the CityScapes dataset specifically, the initial learning rate is adjusted to $3 \times 10^{-4}$. Additionally, a Poly learning rate scheduler with a factor of 1.0 is utilized. For the PASCAL Context and COCO-Stuff datasets, we conduct 80K training iterations, incorporating the same data augmentations as in \cite{mmseg2020}. Training images are resized and cropped to $512 \times 512$ for COCO-Stuff and $480 \times 480$ for PASCAL Context. Single-scale results are reported for model comparisons.

\subsection*{Details of Feed-Forward Network} 
\begin{figure}[h]
  \centering
  \includegraphics[width=\linewidth]{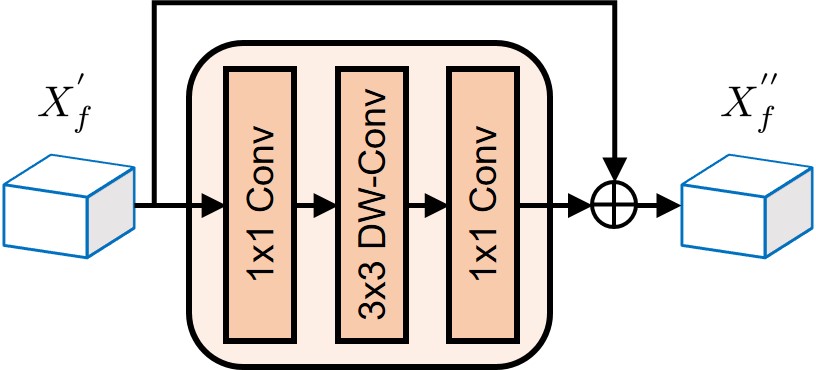}
  \caption{Design of Feed-Forward Network.
  }
  \label{fig:feed_forward_network}
\end{figure}

\vspace{-1mm}
\label{sec:fnn_details}
For the Feed-Forward Network, in the proposed LeMoRe model, we have integrated depth-wise convolution layer between $1 \times 1$ convolution layers and to further minimize the computational complexity, expansion factor of two is incorporated. The design of Feed-Forward Network is shown in~\Cref{fig:feed_forward_network}. Where, let $X_{f}^{'}$ and $X_{f}^{''}$ be the input and output of Feed-Forward Network, respectively.

\subsection*{More Visual Results}
\begin{figure}[h]
  \centering
  \includegraphics[width=\linewidth]{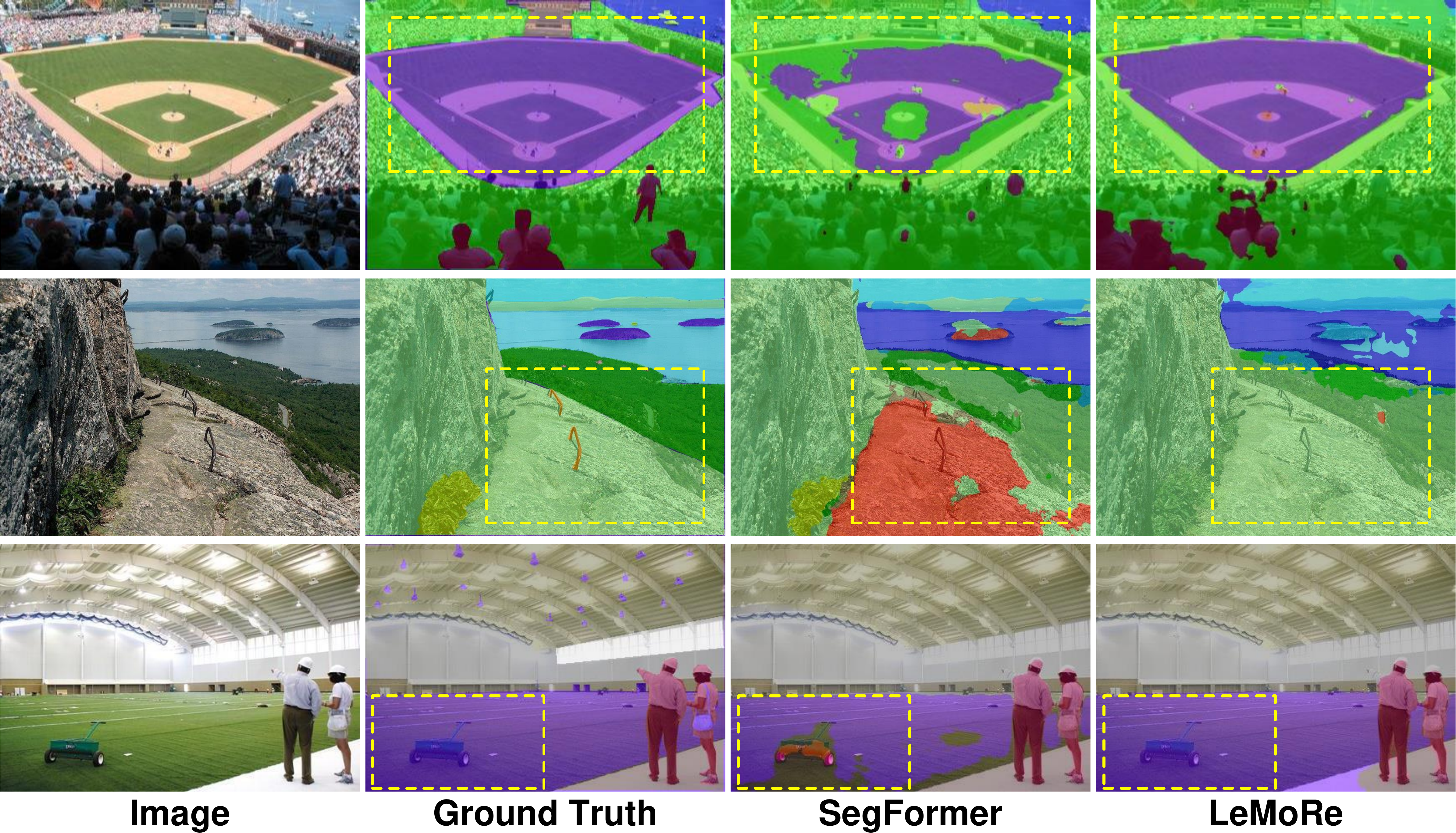}
  \vspace{-6mm}
  \caption{Visualization of Image, Ground Truth, SegFormer, and LeMoRe results on the ADE20K validation set highlights the proposed model's effectiveness in producing high-quality segmentation maps with enhanced spatial consistency.
  }
  \label{fig:Visual_Results_Supplementray}
  \vspace{-4mm}
\end{figure}

More visual results are shown in \Cref{fig:Visual_Results_Supplementray} to demonstrate the efficacy of the proposed approach.


\end{document}